# Fuzzy Approach to Critical Bus Ranking under Normal and Line Outage Contingencies

Shobha Shankar[*], Dr. T. Ananthapadmanabha[**]

[*]Research Scholar and Assistant Professor,
Department of Electrical and Electronics Engineering,
Vidyavardhaka College of Engineering, Mysore, INDIA
Email:shobha.prathi@gmail.com

[**]Professor,
Department of Electrical and Electronics Engineering,
The National Institute of Engineering, Mysore, INDIA
Email:drapn2008@yahoo.co.in

**Abstract.** Identification of critical or weak buses for a given operating condition is an important task in the load dispatch centre. It has become more vital in view of the threat of voltage instability leading to voltage collapse. This paper presents a fuzzy approach for ranking critical buses in a power system under normal and network contingencies based on Line Flow index and voltage profiles at load buses. The Line Flow index determines the maximum load that is possible to be connected to a bus in order to maintain stability before the system reaches its bifurcation point. Line Flow index (LF index) along with voltage profiles at the load buses are represented in Fuzzy Set notation. Further they are evaluated using fuzzy rules to compute Criticality Index. Based on this index, critical buses are ranked. The bus with highest rank is the weakest bus as it can withstand a small amount of load before causing voltage collapse. The proposed method is tested on Five Bus Test System.

## 1 Introduction

Voltage stability and system security are emerging as major problems in the operation of stressed power system. Line outage contingencies are the most common problem in power system and have a considerable effect on altering the base case (pre-contingency case) voltage stability margin of a load bus. Generally, the system continues to operate in the contingency condition for a considerable duration of time, on occurrence of a line outage. The altered voltage stability margins of all the load buses for the various contingency conditions are to be known prior to monitor and initiate emergency control action to avoid voltage collapse. The main cause for voltage collapse is the inability of the system to supply reactive power to cope up with the increasing load growth. The occurrence of voltage collapse is very much dependent upon the maximum load that can be supported at a particular load bus. Any attempt to increase the load beyond this point could force the entire system into instability, leading to voltage collapse. This would indicate that the power system physically could not support the amount of the connected load. Line Flow index (LF index) is used to estimate maximum loadability of a particular load bus in the system. The load buses are ranked according to their maximum loadability, where the load bus having the smallest maximum loadability is ranked highest. Hence this bus is identified as the weakest bus because it can withstand only a small amount of load increase before causing voltage collapse. This information is useful to planning or operation engineers in ensuring that any increment in the system will not exceed the maximum loadability, hence violating the voltage stability limit.

The increase in load of a bus beyond a critical limit pushes the system to the verge of voltage collapse, if the system is not compensated adequately. This critical limit of the bus load is defined as the Voltage Stability Margin. The voltage stability margin estimates the criticality of a bus. Hence, the identification of the critical buses in a system is useful in determining the location of additional voltage support devices to prevent possible voltage instability.

A Fuzzy Set theory based algorithm is used to identify the weak buses in a power system. Bus voltage and reactive power loss at that bus are represented by membership functions for voltage stability study [1]. Newton optimal power flow is used to identify the weakest bus / area, which is likely to cause voltage collapse. The complex power – voltage curve is examined through Newton optimal power flow. The indicator, which identifies the weakest bus, was obtained by integrating all the marginal costs via Kuhn-Tucker theorem [2]. A Fast Voltage Stability Indicator (FVSI) is used to estimate the maximum loadability for identification of weak bus. The indicator is derived from the voltage quadratic equation at the receiving bus in a two bus system. The load of a bus, which is to be ranked is increased till maximum value of FVSI is reached and this load value is used as an indicator for ranking the bus [3]. A weak bus-oriented criterion is used to determine the candidate buses for installing new VAR sources in VAR planning problem. Two indices are used to identify weak buses based on power flow Jacobian matrix calculated at the current operating point of the system [4]. A neural network based method for the identification of voltage-weak buses/areas uses power flow analysis and singular value decomposition method. Kohonen neural network is trained to cluster/rank buses in terms of voltage stability [5]. Voltage Stability Margin Index (VSMI) is developed based on the relationship of voltage stability and angle difference between sending and receiving end buses. VSMI is used to estimate voltage stability margin and identify weak transmission lines and buses at any given operating condition [6]. The weakest bus, most loaded transmission path for that bus from voltage security point of view is identified using nodal voltage security assessment. Control actions are taken to alleviate power flows across that branch to enhance voltage security condition [7]. The Singular Value Decomposition method is used in identifying weak boundaries with respect to voltage instabilities well as in the assessment of the effects of possible disturbances and their corrective actions [8]. The existing techniques [1] - [8] are basically to identify weak buses for a pre contingency system. But for secured operation of the stressed power system, it is essential to know the criticality of a bus at the verge of voltage collapse.

This paper presents a fuzzy approach to rank critical buses in a power system under and normal and network contingencies. Voltage Stability Margin expressed in terms of Static Voltage Collapse Proximity Indicator at critical load of a selected load bus accurately estimates the criticality of that bus from the voltage collapse point of view. Hence the Line Flow index is used as a Static Voltage Collapse Proximity Indicator. The Line Flow index and bus voltage profiles of the load buses are expressed in fuzzy set notation. Further, they are evaluated using fuzzy rules to compute Criticality Index. Critical buses are ranked based on decreasing order of Criticality Index. The proposed approach is tested on Five Bus Test System.

## 2  Formulation of Line Flow Index

Consider a typical transmission line of an interconnected power system shown in Fig.1. Optimal impedance concept used in [9] to develop a simple criterion for voltage stability is as follows;

Load impedance $Z_L \angle \theta$ fed by constant voltage source Vs with internal impedance $Z_S \angle \Phi$ as shown in Fig.2. Application of maximum power transfer theorem to the equivalent circuit shown in Fig.2 results in $Z_L / Z_S = 1$ for maximum power to be flown to the load from the source. $Z_L/Z_S$ is used as the VCPI Voltage Collapse Proximity Indicator. The system is considered to be voltage stable if this ratio is less than 1, other- wise voltage collapse occurs in the system.

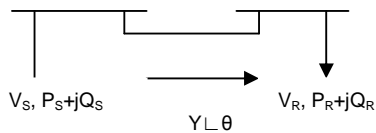

Fig. 1 Typical transmission line of a power system network

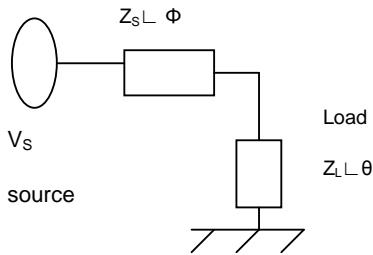

Fig. 2. Two Bus System

The single line model shown in [9] is used, but the system is represented by the admittance model. It is assumed that the load at the bus is total power flow in the represented line. Equivalent admittance model is shown in Fig. 3 where $Y_L \angle \theta$ is the line admittance and the $Y_R \angle \Phi$ is the load admittance and

$\Phi = \tan^{-1}[ Q_r / P_r]$

The indicator is developed with an assumption that only the modulus of the load admittance changes with the change in the system load i.e. it is assumed that always efforts will be made in the system to maintain the constant power factor for the changes in the bus load. Increase in load results in increase in admittance and there by increase in current and the line drop and hence decrease in the voltage at the receiving end.

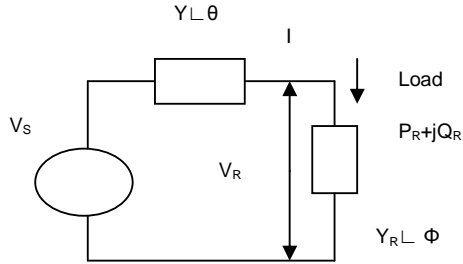

Fig. 3 Transmission line modeled with admittance.

$$I = V_s Y_{eq} \quad\quad\quad (1)$$

$$Y_{eq} = \frac{Y_L Y_R}{\sqrt{Y_L^2 + Y_R^2 + 2Y_L Y_R \cos(\theta - \phi)}}$$

$$V_R = I / Y_R \quad\quad\quad (2)$$

$$= \frac{V_S}{Y_R}\left(\frac{Y_L Y_R}{\sqrt{Y_L^2 + Y_R^2 + 2Y_L Y_R \cos(\theta - \phi)}}\right)$$

Now the active power at the receiving end is given by

$$P_R = V_R I \cos\phi$$

$$P_R = \frac{V_S^2 Y_L^2 Y_R}{\sqrt{Y_L^2 + Y_R^2 + 2Y_L Y_R \cos(\theta - \phi)}} \quad\quad\quad (3)$$

The maximum real power transfer to the bus is obtained by applying the condition $\delta P_R / \delta Y_R = 0$ which leads to a criterion of $|Y_L| = |Y_R|$

Substituting $|Y_L| = |Y_R|$ in equation (3), we get

$$P_{R(max)} = \frac{V_S^2 Y_L \cos\phi}{[2(1 + \cos(\theta - \phi))]} \quad\quad\quad (4)$$

Equation (4) gives the maximum real power that can be transferred through a given line safely without any voltage instability threat. The actual line flow is compared with this maximum power transfer and the stability margin for that line is defined as,

$$LF\ index = \frac{Actual\ real\ power\ flow\ in\ the\ line\ (P_R)}{Maximum\ real\ power\ that\ can\ be\ transferred\ P_{R(max)}}$$

$$LF\ index = \frac{4P_R\ (1+\cos(\theta - \phi))}{V_S^2\ Y_L \cos\phi} \qquad \text{------- (5)}$$

$P_R$ values can be obtained from the load flow solution.

The main cause for the problem of voltage instability leading to voltage collapse is stressed power system characterized by excessive line loading. As the maximum power transfer theory restricts the amount of load that can be transferred through a line, the LF index precisely indicate the voltage stability margin for a selected operating condition.

## 3  Methodology

Voltage Stability Margin expressed in terms of Static Voltage Collapse Proximity Indicator at a given load of a selected load bus accurately estimates the criticality of that bus from the voltage collapse point of view. Hence computation of these indicators along with voltage profiles at load buses can serve as a very good measure in assessing the criticality of a bus. In addition to line flow index, bus voltage profiles are used to identify weak buses under varying load condition. The point at which LF index is close to unity indicates the maximum possible connected load called as maximum loadability at the point of bifurcation. The line flow indices and bus voltage profiles are divided into different categories and are expressed in fuzzy set notation. The severity indices are also divided into different categories. The fuzzy rules are used to evaluate the severity of load buses. Criticality index is computed based on severity of LF index and voltage profiles. Based on this index the buses are ranked. The ranking obtained using Fuzzy approach is verified with Fast Voltage Stability Index (FVSI) [3].

**3.1 Bus voltage profiles**

The bus voltage profiles are divided into three categories using Fuzzy Set notations: low voltage (LV), normal voltage (NV) and over voltage (OV). Figure 4 shows the correspondence between bus voltage profiles and the three linguistic variables.

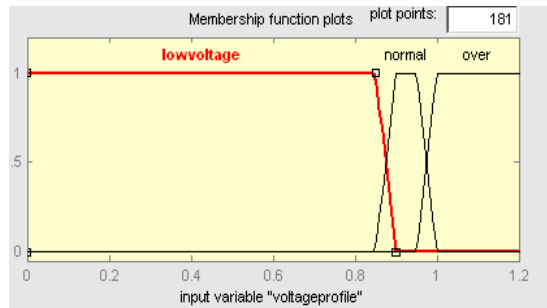

Fig.4 Voltage profiles and the corresponding linguistic variables

**3.2 Line Flow Index**

Fuzzy Approach to Critical Bus Ranking under Normal and Line Outage Contingencies

The Line Flow indices are divided into five categories using Fuzzy Set notations: very small index (VS), small index (S), medium index (M), high index (H), and very high index (VH). Fig. 5 shows the correspondence between the Line Flow index and the five linguistic variables. Fig. 6 shows the severity index for voltage profile and Line Flow index.

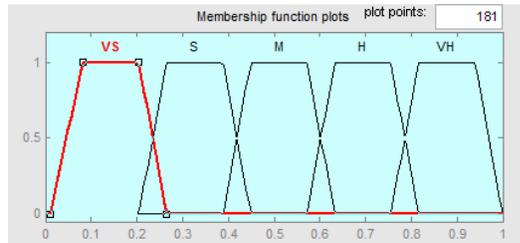

Fig.5: Line flow index and the corresponding linguistic variables

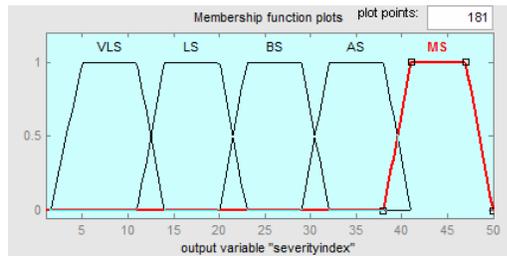

Fig.6 Severity index for voltage profile and line flow index

The fuzzy rules, which are used for evaluation of severity indices of bus voltage profiles and line flow indices, are given in Table 1.

Table 1: Fuzzy rules

| Quantity | Severity |
|---|---|
| Voltage: LV   NV   OV | MS  BS  MS |
| LF index: VS   S   M   H   VH | VLS  LS  BS  AS   MS |

Note: VLS - very less severe; LS - less severe; BS - below severe; AS - above severe; MS - more severe

The Criticality Index is obtained by adding the two severity indices as shown in Fig. 7. The Criticality Index is obtained at critical load for all the load buses. The buses are ranked in decreasing order of Criticality Index.

The following are the steps involved in the approach:

1. Under normal or selected contingency, for a chosen load bus, the reactive power loading is increased until the load flow solution fails to converge. The load prior to divergence is maximum load for that bus.
2. At maximum load, bus voltage profiles and line flow index are determined.
3. Bus voltage profiles and line flow index are expressed in fuzzy set notation.
4. Severity index of line flow index and bus voltage profiles are also represented in fuzzy set notation.
5. Using Fuzzy-If-Then rules severity index for bus voltage profiles and LF index are determined. The FIS is tested in MATLAB 7 Fuzzy Toolbox.
   Criticality index for each load bus is computed using equation,
   $CI = \Sigma SI_{LF} + \Sigma SI_{VP}$
6. The above procedure is repeated for all the load buses and for all critical contingencies.
8. Buses are ranked in decreasing order of Criticality Index.

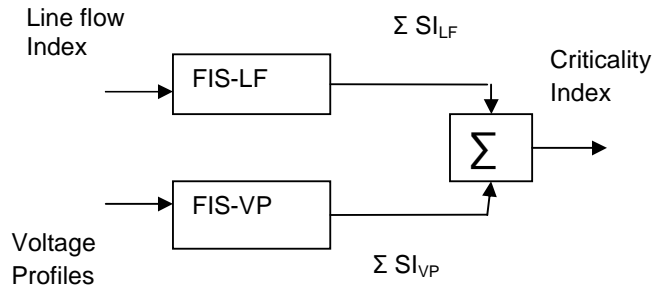

Fig. 7 Parallel Operated Fuzzy Inference System

## 4 Test Results

The proposed approach is tested on Five Bus Test System. It consists of 2 generators, 3 load buses and 7 transmission lines.

A) Without Line Outage Contingencies

Table 2 shows the voltage profile of load buses at base load and critical load at respective load buses. Critical load of a bus is determined by increasing reactive load at that bus until load flow fails to converge. The load prior to convergence is the critical load of that bus. Table 3 shows line flow index at base load and critical load at load buses without line outage contingency. Table 4 and 5 shows severity index for voltage profiles and line flow index calculated using fuzzy rules shown in Table 1. Table 6 provides the criticality index along with rank obtained from fuzzy approach without line outage contingency.

Table-2 Load bus voltage profile at base case load and critical load

| Load bus no. | Voltage in p. u. at base case load | Voltage in p. u. (critical load at bus 3) | Voltage in p. u. (critical load at bus 4) | Voltage in p. u. (critical load at bus 5) |
|---|---|---|---|---|
| 3 | 0.987 | 0.700 | 0.808 | 0.889 |
| 4 | 0.984 | 0.752 | 0.754 | 0.748 |
| 5 | 0.972 | 0.892 | 0.893 | 0.751 |

Table-3 Line flow index for each line at base case load and critical load

| lines | LF index at base case load | LF index critical load at bus 3 | LF index critical load at bus 4 | LF index critical load at bus 5 |
|---|---|---|---|---|
| 1-2 | 0.083 | 0.135 | 0.115 | 0.150 |
| 1-3 | 0.187 | 0.155 | 0.146 | 0.189 |
| 2-3 | 0.128 | 0.092 | 0.075 | 0.061 |
| 2-4 | 0.141 | 0.096 | 0.086 | 0.059 |
| 2-5 | 0.168 | 0.125 | 0.120 | 0.133 |
| 3-4 | 0.015 | 0.325 | 0.015 | 0.006 |
| 4-5 | 0.038 | 0.784 | 0.770 | 0.571 |

Table-4 Severity indices for voltage profiles at critical load

| Load bus no. | Severity Index for voltage profiles($SI_{VP}$) | | |
|---|---|---|---|
|  | Critical load at bus 3 | Critical load at bus 4 | Critical load at bus 5 |
| 3 | 46.2 | 34.8 | 15.4 |
| 4 | 42.3 | 42.7 | 40.2 |
| 5 | 23.5 | 23.0 | 42.0 |
| $\sum SI_{VP}$ | 112 | 100.5 | 97.6 |

Table-5 Severity Indices for LF index at critical load

| Lines | Severity Index for LF index($SI_{LF}$) | | |
|---|---|---|---|
|  | Critical load at bus 3 | Critical load at bus 4 | Critical load at bus 5 |
| 1-2 | 4.71 | 4.71 | 7.61 |
| 1-3 | 8.75 | 6.56 | 12.8 |
| 2-3 | 4.71 | 4.71 | 4.71 |
| 2-4 | 4.71 | 4.71 | 4.71 |
| 2-5 | 4.71 | 4.71 | 4.71 |
| 3-4 | 13.50 | 4.71 | 4.71 |
| 4-5 | 41.90 | 39.60 | 29.40 |
| ∑ $SI_{LF}$ | 82.99 | 69.71 | 68.65 |

Table-6 Criticality Index and Rank using Fuzzy Approach and FVSI

| Bus no. | CI=∑$SI_{VP}$ +∑$SI_{LF}$ | Rank | FVSI | Rank |
|---|---|---|---|---|
| 3 | 194.990 | I | 0.966 | I |
| 4 | 170.210 | II | 0.964 | II |
| 5 | 166.250 | III | 0.679 | III |

From the results, it can be observed that bus number 3 is the most critical bus and bus number 5 is less critical. This indicates that at the verge of voltage instability or voltage collapse, it is the load at bus no. 3 to be monitored and controlled at the earliest. The result obtained from fuzzy approach is compared with Fast Voltage Stability Index (FVSI). The ranking from both the methods agree with each other.

B) Under Critical Line Outage Contingencies

For ranking the critical buses under line outage contingencies, both single line outage and double line outages are considered. On screening, 12 critical line outages are considered. Under these line outages, the load buses are ranked.

Table -7 Bus Ranking at Maximum Load under Critical Contingencies

| Contingency | Bus No. 3 | | Bus No. 4 | | Bus No. 5 | |
|---|---|---|---|---|---|---|
| | Criticality Index | Rank | Criticality Index | Rank | Criticality Index | Rank |
| 1-2 | 108.60 | I | 106 | II | 97.70 | III |
| 2-5 | 119.40 | I | 116.9 | II | 89.60 | III |
| 1-2,2-3 | 108.20 | I | 108.2 | II | 104.00 | III |
| 2-3,2-5 | 114.00 | I | 114 | I | 109.80 | II |
| 2-5,3-4 | 104.00 | I | 95.52 | II | 85.59 | III |
| 1-2,3-4 | 73.58 | III | 96.36 | I | 95.04 | II |
| 2-4,2-5 | 114.00 | I | 114 | I | 104.00 | II |
| 1-2,2-5 | 122.80 | II | 124.4 | I | 121.80 | III |
| 1-2,2-4 | 107.50 | III | 109.6 | II | 112.00 | I |
| 1-3,2-5 | 115.30 | I | 114 | II | 104.00 | III |
| 1-3 | 92.10 | II | 93.8 | I | 80.40 | III |
| 2-4 | 104.00 | I | 101.6 | II | 95.80 | III |

Table -8 Comparison of Critical Bus Ranking using Fuzzy Approach and FVSI Method

| Contingency | Bus No. 3 | | Bus No. 4 | | Bus No. 5 | |
|---|---|---|---|---|---|---|
| | Fuzzy approach | FVSI | Fuzzy approach | FVSI | Fuzzy approach | FVSI |
| 1-2 | I | II | II | I | III | III |
| 2-5 | I | I | II | II | III | III |
| 1-2,2-3 | I | I | II | II | III | III |
| 2-3,2-5 | I | I | I | II | II | III |
| 2-5,3-4 | I | I | II | II | III | III |
| 1-2,3-4 | III | III | I | I | II | II |
| 2-4,2-5 | I | I | I | II | II | III |
| 1-2,2-5 | II | II | I | I | III | III |
| 1-2,2-4 | III | III | II | II | I | I |
| 1-3,2-5 | I | I | II | II | III | III |
| 1-3 | II | II | I | I | III | III |
| 2-4 | I | II | II | I | III | III |

Table 7 gives the Criticality Index and rank for the load buses under selected line outage contingencies. It can be observed that for most of the outages, bus 3 is the most critical bus and bus 5 is less critical. Table 8 provides the comparison of ranking obtained from Fuzzy approach and FVSI method. The rankings obtained from proposed method are very close to the results obtained using FVSI method.

The fuzzy approach effectively ranks the critical buses eliminating the masking effect. The advantage of the proposed method is that it provides the criticality of a bus for a specific contingency whereas the FVSI method identifies the weak bus but does not provide the information about the criticality of a bus with respect to selected contingency. Further, additional voltage support devices can be installed at critical buses to improve system stability.

## Conclusion

A fuzzy based Criticality Index is developed in this paper to rank critical or weak buses in a power system under normal and line outage contingencies. Line Flow index and voltage profiles at load buses are evaluated using fuzzy rules to compute the Criticality Index, which is further used to rank the critical buses. The identification of a critical bus in a power system is useful in determining the location of additional voltage support devices to prevent possible voltage instability. The proposed method is tested on Five Bus Test System.

## Biographies

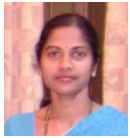

Shobha Shankar received the B.E. degree in Electrical Engineering in 1994, M.Tech degree in Power Systems in 1997 from the University of Mysore, Mysore. She is working as Asst. Professor in the department of Electrical and Electronics Engineering, Vidyavardhaka College of Engineering, Mysore. She is pursing her Doctoral degree from Visvesvaraya Technological University, India in the field of Power Systems.

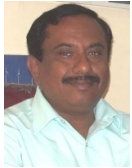

T. Ananthapadmanabha received the B.E. degree in Electrical Engineering in 1980, M.Tech degree in Power Systems in 1984 and Ph.D. degree in 1997 from University of Mysore, Mysore. He is working as Professor, Department of Electrical and Electronics Engineering, The National Institute of Engineering, Mysore. His research interest includes Reactive Power Optimization, Voltage Stability, Distribution Automation and AI applications to Power Systems.